\documentclass[letterpaper]{article} 
\usepackage{aaai24}  
\usepackage{times}  
\usepackage{helvet}  
\usepackage{courier}  
\usepackage[hyphens]{url}  
\usepackage{graphicx} 
\usepackage{natbib}  
\usepackage{caption} 
\usepackage{algorithm}
\usepackage{algorithmic}

\usepackage{newfloat}
\usepackage{listings}
\DeclareCaptionStyle{ruled}{labelfont=normalfont,labelsep=colon,strut=off} 
\floatstyle{ruled}
\newfloat{listing}{tb}{lst}{}
\floatname{listing}{Listing}
\nocopyright

\title{Robust Representation Learning for Unreliable Partial Label Learning}
\author{
    Yu Shi\equalcontrib,
    Dong-Dong Wu\equalcontrib,
    Xin Geng\footnotemark[2],
    Min-Ling Zhang\footnotemark[2]
}

\affiliations{
    School of Computer Science and Engineering, Southeast University, Nanjing 211189, China\\
    \{seushiyu, 220212071, xgeng, zhangml\}@seu.edu.cn
}

\usepackage{algorithm}
\usepackage{algorithmic}
\usepackage{enumitem}
\usepackage{booktabs}
\usepackage{bbding}
\usepackage{multirow}
\usepackage{amsmath}
\usepackage{amssymb}
\usepackage{subcaption}

\begin{document}

\maketitle

\renewcommand{\thefootnote}{\fnsymbol{footnote}}
\footnotetext[2]{Corresponding authors.}

\begin{abstract}
Partial Label Learning (PLL) is a type of weakly supervised learning where each training instance is assigned a set of candidate labels, but only one label is the ground-truth.
However, this idealistic assumption may not always hold due to potential annotation inaccuracies, meaning the ground-truth may not be present in the candidate label set.
This is known as Unreliable Partial Label Learning (UPLL) that introduces an additional complexity due to the inherent unreliability and ambiguity of partial labels, often resulting in a sub-optimal performance with existing methods.
To address this challenge, we propose the Unreliability-Robust Representation Learning framework (URRL) that leverages unreliability-robust contrastive learning to help the model fortify against unreliable partial labels effectively.
Concurrently, we propose a dual strategy that combines KNN-based candidate label set correction and consistency-regularization-based label disambiguation to refine label quality and enhance the ability of representation learning within the URRL framework.
Extensive experiments demonstrate that the proposed method outperforms state-of-the-art PLL methods on various datasets with diverse degrees of unreliability and ambiguity. 
Furthermore, we provide a theoretical analysis of our approach from the perspective of the expectation maximization (EM) algorithm.
Upon acceptance, we pledge to make the code publicly accessible.
\end{abstract}

\section{Introduction}
Partial Label Learning (PLL) typifies a unique form of weakly supervised learning where each training instance is associated with a set of candidate labels, among which only one is the ground-truth. 
Compared to the acquisition process of high-quality annotated datasets, collecting partial labels is both cost-efficient and time-saving.
Such advantages of PLL have garnered significant interest from researchers across various fields, such as web mining \cite{luo2010learning,xuwebly}, multimedia content analysis \cite{zeng2013learning}, automatic image annotations \cite{chen2017learning}, and ecological informatics \cite{LiuD12,briggs2012rank}, etc.

The primary solution for learning from partial labels is disambiguation, which can be categorized into averaging-based and identification-based strategies. Averaging-based disambiguation allows each candidate label to contribute evenly to model training, with the final prediction derived by averaging the outputs of all candidate labels \cite{hullermeier2006learning, cour2011learning,cour2009learning, zhang2015solving}. However, this strategy is susceptible to the adverse influence of false positive labels within the candidate label set.
On the other hand, identification-based disambiguation considers the ground-truth label as a latent variable and iteratively predicts it during optimization. Recent advancements in identification-based disambiguation, particularly using Deep Neural Networks (DNNs), have  yielded impressive performance \cite{jin2002learning, NguyenC08,zhang2016partial}. For example, Lv et al. \cite{lv2020progressive} employed self-training techniques to progressively identify the ground-truth labels during training. Assuming a uniform partial label generation, Feng et al. \cite{feng2020provably} derived two methods which are risk-consistent and classifier-consistent, respectively.
Similarly, Wen et al. \cite{wen2021leveraged} introduced a series of loss functions termed leveraged weighted loss based on the class-conditional partial label generation assumption. 
Wang et al. \cite{wang2022pico} proposed a prototype-based disambiguation mechanism by integrating contrastive representation learning into PLL. Additionally, Wu et al. \cite{wu2022revisiting} revisited consistency regularization in PLL and proposed a regularized training framework that achieves state-of-the-art performance, approximating supervised learning.
Both disambiguation strategies depend on the presumption that the ground-truth label is invariably present within the candidate label set. However, potential errors in annotators' judgment could invalidate this presumption. 

In response to this challenge, Unreliable Partial Label Learning (UPLL) has been proposed \cite{lv2021robustness}, which considers the more realistic scenario wherein the ground-truth label may not be included in the candidate label set. Lv et al. \cite{lv2021robustness} employed noise-tolerant loss functions to align with the ground-truth label, effectively avoiding the interference of noisy labels and other candidate labels. Nevertheless, this approach may not fully utilize the informative features in the representation space and could face limitations when confronted with high levels of unreliability or noise. These observations highlight the necessity for an effective method to address the issue of high unreliability in UPLL.

The fundamental challenge of UPLL lies in managing the unreliable partial labels. A common method for effectively handling unreliable data is to utilize the model's prediction to correct training instances. However, this strategy may induce confirmation bias, leading to sub-optimal performance. To address this issue, we draw inspiration from manifold regularization, which proposes that the manifold structure in the feature space should be preserved in the label space. With this principle as motivation, we propose a novel method that incorporates unreliability-robust contrastive learning. Specifically, our method fosters the learning of robust representations, which effectively counteract the interference of unreliable partial labels, and utilizes information from the feature space to assist in refining the candidate label sets. Our primary contributions can be summarized as follows:

\begin{itemize}[leftmargin=12pt]
\item We propose unreliability-robust contrastive learning, a novel approach to UPLL that mitigates the challenges posed by label unreliability and enhances model robustness. This approach promotes effective learning within an unreliable label space.
\item We introduce a novel strategy that amalgamates a KNN-based candidate label set correction and a consistency regularization based label disambiguation, refining the quality of labels and enhancing representation within the URRL framework. This strategy addresses the challenges posed by ambiguity and unreliability in the label space.
\item Our experimental results demonstrate that our method achieves state-of-the-art performance on multiple datasets with varying degrees of unreliability and ambiguity. Additionally, we provide a theoretical analysis of our method from the perspective of Expectation Maximization algorithm.
\end{itemize}

\section{Related Work}
\textbf{Partial Label Learning} (PLL) grapples with the conundrum where each training instance corresponds to a set of candidate labels, one of which is the ground-truth. This problem, also called ambiguous-label learning \cite{chen2017learning} or superset-label learning \cite{gong2017regularization}, has received much attention recent years \cite{tian2023partial,zeng2013learning,liu2014learnability,gong2022unifying, AGGD-TPAMI}. Existing work can be classified into averaging-based methods and identification-based methods.
\textsl{Averaging-based methods} treat each candidate label equally as a potential ground-truth \cite{hullermeier2006learning, cour2011learning,cour2009learning, zhang2015solving}. However,they are susceptible to false positive labels in the candidate set.
To mitigate this shortcoming, \textsl{identification-based methods} were introduced. They consider the ground-truth label as a hidden variable and iteratively refine its confidence \cite{NguyenC08,LiuD12,zhang2016partial,liu2021partial, feng2019SURE}. They employ machine learning techniques like maximum likelihood \cite{zhou2016partial, liu2012conditional, jin2002learning}, maximum margin \cite{chai2019large, yu2016maximum, lyu2020self}, boosting \cite{tang2017confidence}, and feature-aware disambiguation \cite{zhang2016partial, feng2018leveraging, wang2021learning, feng2019partial, xu2019partial}.

Recently, identification methods combining with deep learning and efficient stochastic optimization, have achieved remarkable performance \cite{wang2022pico,yan2020multi,lyudeep,luoexploring,xuwebly,gong2022partial,cheng2023partial,Qiao2023iclr}.
Yao et al. \cite{yao2020deep} pioneered deep PLL, incorporating temporal ensembling during training. 
Lv et al. \cite{lv2020progressive} devised a rudimentary self-training strategy to gradually identify the ground-truth label. 
Wu et al. \cite{wu2022revisiting} proposed a regularized training framework with consistency regularization to preserve feature and label space manifolds.
In their series of works, Feng et al. \cite{feng2020provably, wen2021leveraged, xu2021instance, Qiao2023iclr} model the generation of partial labels from uniform, class-conditional, and instance-dependent perspectives.
Yan et al. \cite{Yan2023iclrb} were the first to introduce a new problem known as partial label unsupervised domain adaptation, proposing a novel prototype alignment approach in response.

Recently, PLL approaches have been devised to tackle other limited supervision including partial multi-label learning \cite{xie2018partial, cao2021partial, gong2021understanding} and semi-supervised PLL \cite{wang2019partial,wang2020semi, zhang2021exploiting}. Furthermore, several other problem settings have been explored in conjunction with PLL, such as long-tail PLL \cite{hong2023long,wang2022solar,liu2021partial}, multi-view PLL \cite{chen2020multi}, and multi-dimensional PLL \cite{wang2021learning}.
However, these methods assume the ground-truth label is always hidden in the candidate label set. In practice, this may not always hold true. This deviation from the norm, where the ground-truth label does not necessarily hide within the candidate set, is identified as "Unreliable partial label learning" \cite{lv2021robustness}. Although Lv et al. \cite{lv2021robustness} have demonstrated that a bounded loss is robust against unreliability, the accuracy of their proposed Robust Ambiguity Set Based (RABS) model \cite{lv2021robustness} remains limited in the presence of high unreliability.

\textbf{Contrastive learning} is a strategic approach that aims to create a representation space where similar instances are closer together, while dissimilar instances are further apart. This approach has shown significant promise in the field of unsupervised representation learning, as showcased by numerous studies \cite{khosla2020supervised, he2020momentum, chen2020simple, li2020prototypical}. A recent method, termed prototypical contrastive learning (PCL) \cite{li2020prototypical}, utilizes cluster centroids as prototypes, refining the network by bringing an image embedding closer to its assigned prototypes. This success has sparked a series of works that incorporate contrastive learning into weakly supervised learning, which covers partial label learning \cite{wang2022pico, xiaambiguity, he2022partial}, and noisy label learning problems \cite{li2022selective, zheltonozhskii2022contrast}, among others.
Distinct from these methods, our approach introduces a novel dynamic label distribution weighted class prototype, combined with a KNN-based candidate label set correction algorithm, to promote collaboration.

\section{The Proposed Method}
Consider $\mathcal{X}$ and $\mathcal{Y} = \{1, 2, ..., C\}$ as the feature space and $C$-class label space, respectively. An unreliable partial label dataset is denoted as $\mathcal{D}=\{(\boldsymbol{x}_i,\mathcal{S}_i)\}_{i=1}^{n}$, where $\boldsymbol{x}_i$ represents the $i$-th instance and $\mathcal{S}_i$ corresponds to the set of candidate labels. The latent ground-truth label of $\boldsymbol{x}_i$ is denoted as $y_i$. 
Traditionally in partial label learning, the ground-truth label $y_i$ is assumed to be concealed in its candidate label set $\mathcal{S}_i$.
However, when dealing with unreliable partial labeled data, the ground-truth label $y_i$ may not be present in $\mathcal{S}_i$ with some probability.

\begin{figure*}[tb] 
  \centering 
  \includegraphics[width=0.85\textwidth]{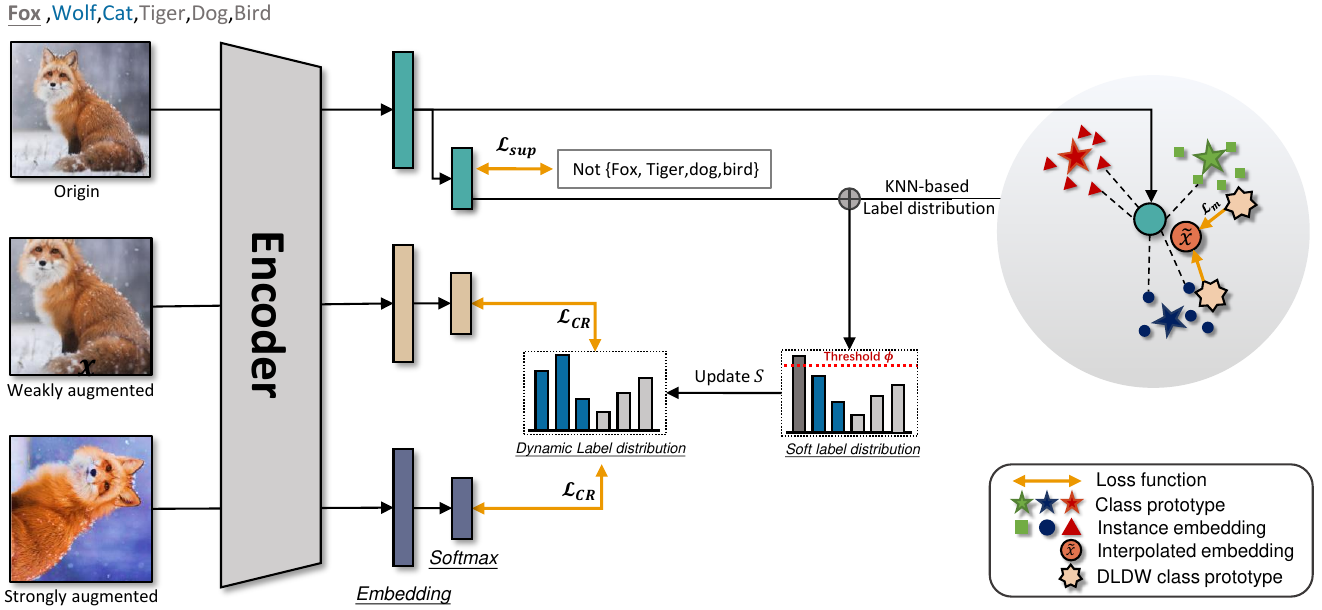} 
  \caption{
  The illustration of our proposed Unreliability-Robust Representation Learning framework. The framework projects images into a lower-dimensional subspace, and the geometric structure of this subspace is regularized using $\mathcal{L}_{\text{m}}$. This term represents a prototypical contrastive loss augmented with mixup, encouraging the embedding of a linearly-interpolated input to maintain a consistent linear relationship with the Dynamic Label Distribution Weighted (DLDW) class prototype. Simultaneously, the softmax prediction of different augmentations is trained to align with the dynamic label distribution $\boldsymbol{l}$, thereby imposing a regularization effect on the classifier. In parallel, the geometric structure of the embeddings coupled with the model prediction is utilized to rectify the candidate label set $\mathcal{S}$.
  } 
  \label{fig:framework} 
\end{figure*}

To address the challenges posed by UPLL, we propose an \textbf{Unreliability-Robust Representation Learning} framework (\textbf{URRL}), which involves two iterative steps: 1) unreliability-robust contrastive learning trains the network to acquire robust representations, and 2) progressive label refinement gradually disambiguates candidate label sets and adds potential ground-truth labels to their corresponding sets.
The entire framework is illustrated in Fig. \ref{fig:framework}, and the comprehensive procedure is presented in Algorithm \ref{alg:URRL}.

\subsection{Unreliability-Robust Contrastive Learning}
The key challenges in UPLL arise from the ambiguity and unreliability, which hinder the acquisition of robust representations. 
Prior to learn robust representations, optimizing the Cross-Entropy loss directly will result in error accumulation due to the presence of false positive and false negative labels.
However, recent studies \cite{gao2021discriminative, wu2022revisiting} have demonstrated the efficacy of optimizing on complementary labels, which motivates us to focus on non-candidate labels. 
As non-candiate labels mainly contain false negative labels in UPLL, minimizing the negative log-likelihood of outputs of these labels would provide a reliable basis for learning in the context of UPLL.
Therefore, the supervised loss is formulated as the following negative log-likelihood loss
\begin{equation}
\label{eq:sup}
\mathcal{L}_{\text{Sup}}(\boldsymbol{x}_i, \mathcal{S}_i)=-\sum_{j \notin \mathcal{S}_i} \log \left(1-f_j(\boldsymbol{q}_i)\right),
\end{equation}
where $\boldsymbol{q}_i=g(\boldsymbol{x}_i)$ means that the encoder network $g(\cdot)$ maps $\boldsymbol{x}_i$ to a latent representation $\boldsymbol{q}_i$ and $f(\cdot)$ is the classifier making the final prediction.

After obtaining the supervised loss, our goal is to incorporate the structural knowledge of classes into the embedding space. 
Nevertheless, a critical question arises: ``How can we obtain the prototypes of classes from an ambiguous and unreliable label space?''. 
To address this issue, we propose the dynamic label distribution denoted by $\boldsymbol{l}_i$ with respect to $\boldsymbol{x}_i$. 
Specifically, the dynamic label distribution refines the given candidate set by utilizing the probability and robust representation, and it is updated at the end  of each epoch during the training process.
We define the normalized class prototype as follows:
\begin{equation}
    \boldsymbol{z}_{c}=\texttt{Normalize}\left(\frac{1}{n}\sum_{i=1}^{n} \boldsymbol{q}_{i}\!\cdot\! \boldsymbol{l}_{i}^{c}\right),
    \label{Eq.proto}
\end{equation}
where $\boldsymbol{q}_{i}$ is the embedding of the $i$-th image, while $\boldsymbol{l}_{i}^{c}$ represents the probability of the $c$-th class in its dynamic label distribution.
To improve the similarity between the image embedding $\boldsymbol{q}_i$ and its corresponding dynamic-label-distribution-weighted class prototype $\boldsymbol{p}_i$, which is expressed as $\sum_{c=1}^{C}{\boldsymbol{z}_{c} \!\cdot\!\boldsymbol{l}_{i}^{c}}$, and to make it stand out from other class prototypes, we define the following prototypical contrastive loss as follows:
\begin{equation}
    {\mathcal L}_{\text{pc}}(\boldsymbol{q}_i, \boldsymbol{l}_i)=-\log\frac{\exp(\boldsymbol{q}_i\!\cdot\!\boldsymbol{p}_i / \tau)}
    {\sum_{c=1}^{C}\exp(\boldsymbol{q}_i\!\cdot\!\boldsymbol{z}_{c}/\tau)},
\end{equation}
where $\tau$ represents the temperature parameter.
Since the dynamic label distribution $\boldsymbol{l}_i$ can be effected by unreliability and ambiguity, we would like to regularize the encoder from memorizing unreliable labels. 
Drawing inspiration from the mixup strategy \cite{zhang2017mixup}, which has proven effective in tackling label noise, we adopt a similar strategy within our framework.
We generate virtual training instances through the linear interpolation of a given instance with another instance randomly selected from the same batch. 
Specifically, for $i$-th image, we denote the virtual training instances $\Tilde{\boldsymbol{x}}_{i} = \lambda \boldsymbol{x}_{i} + (1-\lambda) \boldsymbol{x}_{j}$, where $\boldsymbol{x}_{j}$ is randomly chosen from the same batch, and $\lambda\!\sim\!\text{Beta}(\alpha, \alpha)$.
By setting $\Tilde{\boldsymbol{q}}_i$ as the embedding of $\Tilde{\boldsymbol{x}}_i$ and $\boldsymbol{l}_j$ as the dynamic label distribution of $\boldsymbol{x}_{j}$ respectively, we define the mixup version of the prototypical contrastive loss as a weighted combination formulation:
\begin{equation}
    {\mathcal{L}}_{\text{m}}(\boldsymbol{q}_i,\boldsymbol{l}_i)=\lambda{\mathcal{L}}_{\text{pc}}( \Tilde{\boldsymbol{q}_i}, \boldsymbol{l}_i )+(1-\lambda){\mathcal{L}}_{\text{pc}}( \Tilde{{\boldsymbol{q}}}_i, \boldsymbol{l}_j ).
\end{equation}
By incorporating the mixup strategy into the prototypical contrastive loss, the embedding for the interpolated input is able to maintain a consistent linear relationship with the dynamic-label-distribution-weighted class prototypes, helping to further enhance the model's ability to learn robust representation.

In addition to utilizing prototypical contrastive learning to explicitly learn robust representations of the model in feature space based on prototype features, consistency regularization is also an implicit feature alignment method. It enhances the robustness of the model in the presence of noise by preserving consistency in the output of instances with different data augmentation in the label space.
To achieve this, the output distribution of weak and strong augmentations for instance $\boldsymbol{x}$ is aligned by minimizing the divergence between each output pair derived from the weakly-augmented instance $\boldsymbol{x}_i^w$ and the strongly-augmented instance $\boldsymbol{x}_i^s$. 
However, pairwise divergence minimization may lead to significant semantic shifts that could compromise model performance \cite{wu2022revisiting}. To circumvent this challenge, we propose a novel alignment between different augmentations $\{\boldsymbol{x}_i^w, \boldsymbol{x}_i^s\}$ and the dynamic label distribution $\boldsymbol{l}_i$ with respect to $\boldsymbol{x}_i$. Specifically, the consistency regularization term is given by
\begin{equation}
\mathcal{L}_{\text{CR}}(\boldsymbol{x}_i, \boldsymbol{l}_i)=- (\texttt{KL}( \boldsymbol{l}_i \| f(g(\boldsymbol{x}^w_i))) + \texttt{KL}( \boldsymbol{l}_i \| f(g(\boldsymbol{x}_i^s)))),
\end{equation}
where $\texttt{KL}(\cdot||\cdot)$ denotes the Kullback-Leibler divergence.
Consequently, the comprehensive training objective is designed to minimize the cumulative sum of all loss functions, as formally expressed as
\begin{equation}
\mathcal{L}= \mathcal{L}_{\text {Sup}} + w_{\text{m}}\mathcal{L}_{\text {m}} + w_{\text{CR}}\mathcal{L}_{\text {CR}},
\label{Eq.loss}
\end{equation}
where both $w_{\text{m}}$ and $w_{\text{CR}}$ are the weight factor.

\begin{algorithm}[tb]
    \caption{Pseudo-code for Proposed Method URRL}
    \label{alg:URRL}
    \textbf{Input}: The classifier $f$ with parameters $\boldsymbol{\theta}$; unreliable partial label dataset $\mathcal{D}$; training epochs $T$ and iterations $I$.   \\
    \textbf{Output}: Learned parameters $\boldsymbol{\theta}$ for classifier $f$ and encoder $g$.

    \begin{algorithmic}[1] 
    \STATE Randomly initialize $\boldsymbol{\theta}$
    \FOR{$t \leftarrow 1$ to $T$}
        \STATE Calculate prototypes $\boldsymbol{z}$ by Eq. (\ref{Eq.proto})
        \STATE Candidate label set correction according Eq. (\ref{Eq.PLR_1})
        \FOR{$i \leftarrow 1$ to $I$}
            \STATE Calculate loss according Eq. (\ref{Eq.loss}) and update $\boldsymbol{\theta}$
            \STATE Get updated dynamic label distribution $\boldsymbol{l}$ via Eq. (\ref{Eq.PLR_2})
        \ENDFOR
    \ENDFOR
    \end{algorithmic}
\end{algorithm}

\subsection{Progressively Label Refinement}

\subsubsection{KNN-Based Candidate Label Set Correction}
As described, the unreliability-robust contrastive loss poses a clustering effect in the embedding space. Collaboratively, we then introduce a novel $K$NN-based candidate label set correction strategy.
Specifically, we calculate the soft label distribution according to
\begin{multline}
    \boldsymbol{\pi}_{i}=\frac{1}{2}f(\boldsymbol{x}_{i})+\frac{1}{2}\sum_{j=1}^{K}w_{ij}\boldsymbol{l}_{j}, \\
    \text { where } w_{ij}=\frac{\exp(\boldsymbol{q}_{i}\!\cdot\!\boldsymbol{q}_{j} / \tau)}{ \sum_{j=1}^{K} \exp( \boldsymbol{q}_{i}\!\cdot\!\boldsymbol{q}_{j}/\tau ) }.
\end{multline}

In an effort to refine the candidate label set, we leverage pseudo labels with a goal of including classes that exhibit high probability in the pseudo labels. Consequently, the candidate label set $\mathcal{S}_i$ is dynamically updated at each epoch according to the following formulation:

\begin{equation}
\mathcal{S}_{i}=\left\{\begin{array}{cc}
\mathcal{S}_{i} \cup \{c\} & \text { if } c \notin \mathcal{S}_{i} \text{ and } \boldsymbol{\pi}_{i}^{c} > \phi; \\
\mathcal{S}_{i} & \text {otherwise}.
\end{array}\right.
\label{Eq.PLR_1}
\end{equation}

In this subsection, $c=\operatorname{argmax}({ \boldsymbol{\pi}_{i} })$, and $\phi$ represents the update threshold. Essentially, if the probability of a specific class $c$ surpasses the threshold $\phi$ but is not currently included in the candidate label set $\mathcal{S}_i$, the class $c$ will be incorporated into $\mathcal{S}_i$.

\subsubsection{Consistency-Regularization-Based Label Disambiguation}
Ambiguity within the label space will detrimentally affect model training. Hence, the initial step in label refinement involves label disambiguation. 
Fortunately, Wu et al. \cite{wu2022revisiting} have demonstrated the effectiveness of the Lagrangian multiplier method, premised on consistency regularization, in facilitating the explicit disambiguation of candidate labels.
By utilizing this technique, the dynamic label distribution can be updated as follows:
\begin{equation}
    \boldsymbol{l}_i^{c}(\boldsymbol{x}_i)=\frac{f_{c}^{\frac{1}{2}}(g(\boldsymbol{x}_i^w)) \cdot f_{c}^{\frac{1}{2}}(g(\boldsymbol{x}_i^s))} {\sum_{j\in \mathcal{S}_{i}}\left(f_{j}^{\frac{1}{2}}(g(\boldsymbol{x}_i^w)) \cdot f_{j}^{\frac{1}{2}}(g(\boldsymbol{x}_i^s))\right)}.
    \label{Eq.PLR_2}
\end{equation}

Unreliability-Robust Contrastive Learning enhances the quality of representations and synergistically interacts with Progressive Label Refinement. This can be logically reasoned as a process of maximizing likelihood via clustering similar instances, a concept rooted in the expectation-maximization perspective. The theoretical analysis on the EM perspective is elaborated in the Section \ref{sec.5}.

\section{Experiments}
\label{experiments}

In this section, we present a detailed account of the experimental procedures undertaken to evaluate the efficacy of the proposed method. Our objective is to showcase the robustness of the proposed approach by comparing its performance with that of seven state-of-the-art algorithms under varying levels of ambiguity and unreliable rate. By doing so, we aim to establish the superiority of our method over existing techniques.

\subsection{Experiment setup}
\textbf{Datasets.}
In this research, we scrutinize the performance of our proposed method by leveraging two artificially synthesized benchmarks: CIFAR-10 and CIFAR-100 \cite{cifar}. 
In addition, we further extend our evaluations to the Fashion-MNIST dataset \cite{fashionMNIST}. Detailed results and a comprehensive analysis pertaining to Fashion-MNIST dataset are presented in Appendix B.2.
These datasets undergo a label corruption process, followed by partial label generation via a flipping operation, adhering to the strategies stipulated in \cite{lv2021robustness}.
In the context of our study, we introduce two key parameters. We denote the rate of unreliable label generation as $\mu$, and the partial rate as $\eta$.
In this process, for a given instance $\boldsymbol{x}_{i}$ with an initial ground-truth label $y=i, i \in [1, C]$, a fixed probability $1-\mu$ is assigned to maintain the label as is. 
Conversely, there exists a probability $\mu / (C-1)$ of the label flipping to $j$, where $j \in \mathcal{Y}, j \neq i$. 
Subsequently, we instigate the transition of negative labels to false positive labels with a probability $\eta$. 
In essence, all $C-1$ negative labels possess an equal opportunity of transforming into false positives. 
The aggregated set, comprising of the ground-truth and flipped labels, forms the candidate set $\mathcal{S}$.
For experimental analysis, the datasets are systematically partitioned into training, validation, and test sets in a 4:1:1 ratio.
\begin{table*}[tbp]
    \centering
    \resizebox{0.9\linewidth}{!}{
        \begin{tabular}{c|c|c|ccccccc}
            \midrule
            Dataset & \multicolumn{1}{c}{$\eta$} 
                & $\mu$         & Ours                        & RABS               & PiCO               & CR-DPLL            & PRODEN             & RC                 & CC                  \\
            \midrule
            \multirow{9}[6]{*}{CIFAR-10}
                & 0.1   & 0.1   & \textbf{92.26 $\pm$ 0.53\%} & 79.11 $\pm$ 0.93\% & 90.32 $\pm$ 0.28\% & 89.70 $\pm$ 0.18\% & 77.89 $\pm$ 0.61\% & 76.42 $\pm$ 0.28\% & 77.57 $\pm$ 0.51\%  \\
                & 0.1   & 0.2   & \textbf{92.58 $\pm$ 0.43\%} & 76.45 $\pm$ 0.74\% & 87.97 $\pm$ 0.14\% & 86.93 $\pm$ 0.46\% & 71.38 $\pm$ 0.36\% & 69.23 $\pm$ 0.58\% & 70.21 $\pm$ 0.38\%  \\
                & 0.1   & 0.3   & \textbf{91.87 $\pm$ 1.12\%} & 71.07 $\pm$ 0.96\% & 85.80 $\pm$ 0.39\% & 84.92 $\pm$ 0.53\% & 63.73 $\pm$ 0.52\% & 62.11 $\pm$ 0.93\% & 62.39 $\pm$ 0.64\%  \\
                \cmidrule{2-10}      
                & 0.3   & 0.1   & \textbf{91.81 $\pm$ 0.66\%} & 45.33 $\pm$ 1.96\% & 89.77 $\pm$ 0.34\% & 89.23 $\pm$ 0.34\% & 73.96 $\pm$ 0.69\% & 69.53 $\pm$ 0.36\% & 73.10 $\pm$ 0.26\%  \\
                & 0.3   & 0.2   & \textbf{91.88 $\pm$ 0.71\%} & 41.04 $\pm$ 0.29\% & 87.43 $\pm$ 0.32\% & 85.64 $\pm$ 0.18\% & 66.15 $\pm$ 0.52\% & 62.35 $\pm$ 0.56\% & 64.93 $\pm$ 1.37\%  \\
                & 0.3   & 0.3   & \textbf{91.68 $\pm$ 0.45\%} & 38.53 $\pm$ 1.66\% & 84.66 $\pm$ 0.27\% & 82.88 $\pm$ 0.18\% & 57.92 $\pm$ 0.47\% & 55.67 $\pm$ 0.25\% & 56.62 $\pm$ 0.25\%  \\
                \cmidrule{2-10}      
                & 0.5   & 0.1   & \textbf{92.67 $\pm$ 0.58\%} & 27.55 $\pm$ 0.90\% & 88.67 $\pm$ 0.32\% & 89.01 $\pm$ 0.60\% & 63.52 $\pm$ 0.28\% & 56.54 $\pm$ 0.26\% & 68.01 $\pm$ 0.45\%  \\
                & 0.5   & 0.2   & \textbf{92.55 $\pm$ 0.33\%} & 26.29 $\pm$ 2.18\% & 85.83 $\pm$ 0.04\% & 85.00 $\pm$ 0.33\% & 56.30 $\pm$ 0.36\% & 51.90 $\pm$ 0.50\% & 58.40 $\pm$ 0.71\%  \\
                & 0.5   & 0.3   & \textbf{92.69 $\pm$ 0.56\%} & 21.65 $\pm$ 1.18\% & 82.31 $\pm$ 0.59\% & 80.16 $\pm$ 0.49\% & 51.69 $\pm$ 0.36\% & 47.02 $\pm$ 0.80\% & 50.01 $\pm$ 0.09\%  \\
            \midrule
            \midrule

            \multirow{9}[6]{*}{CIFAR-100} 
                & 0.01  & 0.1   & \textbf{68.21 $\pm$ 0.03\%} & 27.63 $\pm$ 0.48\% & 65.90 $\pm$ 0.16\% & 64.96 $\pm$ 0.07\% & 48.66 $\pm$ 0.77\% & 45.87 $\pm$ 0.36\% & 47.22 $\pm$ 0.86\%  \\
                & 0.01  & 0.2   & \textbf{65.84 $\pm$ 0.68\%} & 22.50 $\pm$ 0.84\% & 61.74 $\pm$ 0.83\% & 60.71 $\pm$ 0.63\% & 41.31 $\pm$ 1.02\% & 38.28 $\pm$ 0.48\% & 39.80 $\pm$ 0.13\%  \\
                & 0.01  & 0.3   & \textbf{63.83 $\pm$ 0.42\%} & 18.78 $\pm$ 1.43\% & 59.20 $\pm$ 0.29\% & 58.47 $\pm$ 0.37\% & 33.60 $\pm$ 0.16\% & 31.08 $\pm$ 0.76\% & 33.25 $\pm$ 0.61\%  \\
                \cmidrule{2-10}      
                & 0.05  & 0.1   & \textbf{67.65 $\pm$ 0.47\%} & 23.52 $\pm$ 1.85\% & 64.45 $\pm$ 0.26\% & 64.43 $\pm$ 0.55\% & 42.21 $\pm$ 0.49\% & 32.27 $\pm$ 0.19\% & 43.15 $\pm$ 0.48\%  \\
                & 0.05  & 0.2   & \textbf{65.65 $\pm$ 0.56\%} & 17.66 $\pm$ 1.64\% & 61.10 $\pm$ 0.18\% & 60.21 $\pm$ 0.17\% & 33.07 $\pm$ 0.95\% & 29.94 $\pm$ 0.71\% & 35.55 $\pm$ 0.39\%  \\
                & 0.05  & 0.3   & \textbf{62.97 $\pm$ 0.28\%} & 15.68 $\pm$ 0.56\% & 55.81 $\pm$ 0.51\% & 56.59 $\pm$ 0.64\% & 26.41 $\pm$ 0.97\% & 22.65 $\pm$ 0.33\% & 29.34 $\pm$ 0.76\%  \\
            \cmidrule{2-10}      
                & 0.1   & 0.1   & \textbf{67.84 $\pm$ 0.25\%} & 12.78 $\pm$ 0.73\% & 53.05 $\pm$ 3.33\% & 64.08 $\pm$ 0.33\% & 32.84 $\pm$ 0.86\% & 24.17 $\pm$ 1.06\% & 39.22 $\pm$ 0.34\%  \\
                & 0.1   & 0.2   & \textbf{65.42 $\pm$ 0.27\%} & 11.98 $\pm$ 0.22\% & 48.22 $\pm$ 1.82\% & 59.41 $\pm$ 0.28\% & 25.30 $\pm$ 0.43\% & 19.05 $\pm$ 0.28\% & 31.62 $\pm$ 0.11\%  \\
                & 0.1   & 0.3   & \textbf{62.08 $\pm$ 0.78\%} &  9.45 $\pm$ 0.62\% & 41.44 $\pm$ 1.48\% & 53.81 $\pm$ 0.65\% & 19.42 $\pm$ 0.46\% & 15.12 $\pm$ 0.34\% & 23.97 $\pm$ 0.81\%  \\
            \midrule
            \midrule
            \end{tabular}%
        }
    \caption{Test accuracy (mean$\pm$std) on the CIFAR-10 and CIFAR-100 synthesized dataset. Results demonstrating superior performance are distinguished by boldface.}
    \label{tab:cifar10cifar100}
\end{table*}

\textbf{Implementation Details.}
In our investigation of unreliable partial label learning, we employ the PreAct ResNet-18 \cite{he2016identity} as our predictive model. The learning rate is set to $5e-2$, and the weight decay parameter is $1e-3$. We assign  $w_{\text{m}}=5$, $w_{\text{CR}}=1$ and $\tau=0.3$ for the corresponding weights. For the neighbors count of KNN, we set $K=200$. The parameter $\phi$ is configured as $0.7$ for CIFAR-10 and $0.8$ for CIFAR-100 datasets.
We adhere to the data augmentation techniques described in the prior work on CR-DPLL \cite{wu2022revisiting}. To ensure a fair comparison, we adopt the 'strong augmentation' strategy employed in the compared methods, such as RABS, PRODEN, RC, and CC. For PiCO and CR-DPLL, we follow the augmentation settings recommended in their respective literature. Additional implementation details can be found in Appendix B.1.

The empirical validation of our proposed method is carried out using Stochastic Gradient Descent (SGD) \cite{robbins1951stochastic} as the optimization algorithm, with a set momentum of $0.9$. A cosine learning rate decay \cite{loshchilov2016sgdr} is utilized for learning rate scheduling. Each model is trained with a maximum of 500 epochs and incorporated an early stopping strategy with a patience of 25 epochs. In this context, if there is no improvement in accuracy on the validation set for 25 consecutive epochs, the training process is halted. The computational experiments are conducted on NVIDIA RTX 3090 and Tesla V100 GPUs. Moreover, the method's implementation is realized using the PyTorch \cite{paszke_pytorch_2019} framework. We present the final performance by detailing the test accuracy corresponding to the best accuracy achieved on the validation set for each run. Results are reported as mean and standard deviation values, based on three independent runs with different random seeds.

\textbf{Compared Methods.}
To substantiate the effectiveness of our proposed technique and to delve into its inherent attributes, we undertake comparative analyses against seven benchmark methodologies. These comprise of one Unreliable Partial Label Learning (UPLL) method and five advanced Partial Label Learning (PLL) approaches:
1) RABS \cite{lv2021robustness}: This represents an unreliable PLL technique which has demonstrated the robustness of the Average-Based Strategy (ABS) when paired with a bounded loss function, thereby mitigating the impact of unreliability. In our experimental setup, we selected the Mean Average Error (MAE) loss as the baseline.
2) PiCO \cite{wang2022pico}: This PLL method amalgamates the concept of contrastive learning with a class prototype-based label disambiguation technique.
3) CR-DPLL \cite{wu2022revisiting}: This represents a deep PLL method that is predicated on consistency regularization.
4) PRODEN \cite{lv2020progressive}: This PLL approach progressively identifies true labels within candidate label sets.
5) RC \cite{feng2020provably}: This technique represents a risk-consistent method for PLL, employing an importance re-weighting strategy.
6) CC \cite{feng2020provably}: This is a classifier-consistent method for PLL, utilizing a transition matrix to formulate an empirical risk estimator.

\subsection{Experiment Results}
Table \ref{tab:cifar10cifar100} reports the experimental results on CIFAR-10 and CIFAR-100 synthesized datasets. 
An examination of the results obtained from our experiments on the CIFAR-10 and CIFAR-100 datasets, presented in Table \ref{tab:cifar10cifar100}, clearly underlines the superior performance of our proposed Unreliability-Robust Representation Learning (URRL) method. Regardless of the changes in the parameters $\eta$ and $\mu$, our method consistently outperformed the other comparative methods: RABS, PiCO, CR-DPLL, PRODEN, RC, and CC.
In the CIFAR-10 experiment, the URRL method's test accuracy ranged from 91.68\% to 92.69\%, significantly outperforming the other models across all parameter combinations. Remarkably, even under the most challenging conditions ($\eta = 0.5, \mu = 0.3$), our method still achieved an impressive accuracy of 92.69\%.
Similarly, in the CIFAR-100 experiment, our proposed method is distinctly superior, achieving test accuracies between 62.08\% and 68.21\%. Even under the most challenging conditions ($\eta = 0.1, \mu = 0.3$), the URRL method yielded a test accuracy of 62.08\%, demonstrating its robustness and adaptability.
These findings underscore the ability of the URRL method to effectively cope with unreliability in unreliable partial label learning, and its capacity to consistently deliver superior performance across various scenarios and datasets.
We also provide a comprehensive T-distributed Stochastic Neighbor Embedding (t-SNE) analysis to visualize the high-dimensional data, the details of which are elaborated in Appendix B.3.

\textbf{Ablation Study.}
\begin{table}[t]
    \centering
    \resizebox{0.95\columnwidth}{!}{
        \begin{tabular}{c|ccc|c|c}
        \toprule
        Ablation & \ \ \ $\mathcal{L}_{\text{m}}$\ \ \ \   & $\mathcal{L}_{\text{CR}}$ \ \ \  & $\mathcal{S}$-Corr. & CIFAR-10 & CIFAR-100  \\
        \midrule
        URRL  & \Checkmark & \Checkmark & \Checkmark & \textbf{91.68 $\pm$ 0.45\%} & \textbf{62.97 $\pm$ 0.28\%} \\
        w/o $\mathcal{S}$-Corr. & \Checkmark & \Checkmark & \XSolid & 88.96 $\pm$ 0.39\% & 62.47 $\pm$ 0.57\% \\
        w/o $\mathcal{S}$-Corr. \& $\mathcal{L}_{\text{CR}}$ & \Checkmark & \XSolid & \XSolid & 81.50 $\pm$ 0.59\% & 54.58 $\pm$ 0.62\% \\
        w/o $\mathcal{S}$-Corr. \& $\mathcal{L}_{\text{CR}}$ \& $\mathcal{L}_{\text{m}}$ & \XSolid & \XSolid & \XSolid & 75.65 $\pm$ 0.64\% & 41.66 $\pm$ 0.23\% \\
        \bottomrule
        \bottomrule
        \end{tabular}%
    }
    \caption{Ablation study on CIFAR-10 ($\eta=0.3$, $\mu=0.3$) and CIFAR-100 ($\eta=0.05$, $\mu=0.3$).}
    \label{tab:ablation_tab}
\end{table}
The ablation study presented in Table \ref{tab:ablation_tab} provides insights into the contributions of different components of the URRL method. The complete URRL model, which includes $\mathcal{L}_{\text{m}}$, $\mathcal{L}_{\text{CR}}$, and the KNN-based correction module, achieved the best performance on both CIFAR-10 (91.68\% $\pm$ 0.45\%) and CIFAR-100 (62.97\% $\pm$ 0.28\%) under the specified parameters. When we removed the KNN-based correction module $\mathcal{S}$ from URRL, there is a noticeable decrease in performance. The accuracy dropped to 88.96\% $\pm$ 0.39\% on CIFAR-10 and 62.47\% $\pm$ 0.57\% on CIFAR-100, underscoring the importance of $\mathcal{S}$-Correction in boosting the model's performance. The role of $\mathcal{L}_{\text{CR}}$ becomes clear when we compare the performance of URRL without the KNN-based $\mathcal{S}$-correction and $\mathcal{L}_{\text{CR}}$. There is a significant drop in accuracy to 81.50\% $\pm$ 0.59\% on CIFAR-10 and 54.58\% $\pm$ 0.62\% on CIFAR-100. This highlights the key role of $\mathcal{L}_{\text{CR}}$ in improving model performance by correcting labels based on the dynamic label distribution.
Finally, when all additional components ($\mathcal{L}_{\text{m}}$, $\mathcal{L}_{\text{CR}}$, and $\mathcal{S}$-Correction) are removed, the performance dropped dramatically to 75.65\% $\pm$ 0.64\% on CIFAR-10 and 41.66\% $\pm$ 0.23\% on CIFAR-100, demonstrating the crucial contributions of these components to the URRL method. This result further highlights the strength of our full URRL model, which utilizes all these components to effectively tackle the unreliable partial label learning problem.

\textbf{Analysis on Parameter $\phi$.}
The importance of the $\phi$ parameter in our URRL method can be observed through the Table \ref{tab:phi}. As the value of $\phi$ changes, the performance on both CIFAR-10 and CIFAR-100 datasets demonstrates notable variation. For the CIFAR-10 dataset ($\eta=0.3$, $\mu=0.3$), the highest accuracy of 91.68\% (± 0.45\%) is achieved with a $\phi$ value of 0.7.

\begin{table}[tb]
    \centering
    \resizebox{0.55\columnwidth}{!}{
    \begin{tabular}{c|c|c}
    \toprule
    $\phi$  & CIFAR-10 & CIFAR-100 \\
    \midrule
    0.5   & 82.46 $\pm$ 2.41\% & 45.48 $\pm$ 0.86\% \\
    0.6   & 90.25 $\pm$ 0.54\% & 52.31 $\pm$ 0.47\% \\
    0.7   & \textbf{91.68 $\pm$ 0.45\%} & 59.51 $\pm$ 0.07\% \\
    0.8   & 89.41 $\pm$ 0.35\% & \textbf{62.97 $\pm$ 0.28\%} \\
    0.9   & 88.91 $\pm$ 0.52\% & 62.54 $\pm$ 0.18\% \\
    \bottomrule
    \bottomrule
    \end{tabular}
    }
    \caption{Accuracy comparison with different $\phi$.}
    \label{tab:phi}
\end{table}

Similarly, on the CIFAR-100 dataset ($\eta=0.05$, $\mu=0.3$), 
the maximum accuracy of 62.97\% (± 0.28\%) is obtained when $\phi$ is set to 0.8. The results indicate that the selection of an appropriate $\phi$ value is crucial for the performance of the URRL method.
It is worth noting that, as the value of $\phi$ increases from 0.5 to 0.7 for CIFAR-10 and from 0.5 to 0.8 for CIFAR-100, the accuracy improves considerably. However, when $\phi$ exceeds these optimal values, the performance tends to decline. In our experiments, we set $\phi=0.7$ and $\phi=0.8$ for CIFAR-10 and CIFAR-100 respectively.

\section{Theoretical Understandings}
\label{sec.5}
\noindent \textbf{Overview.}
Our learning objective follows the Expectation-Maximization (EM) algorithm.
In the \textbf{E-step}, each instance $\boldsymbol{x}_i\!\in\!\mathcal{D}$ is assigned to one specific cluster, which is estimated by using the one-hot prediction $\tilde{\boldsymbol{y}}_i\!=\!\texttt{Onehot}(\operatorname{argmax}_{j \in \mathcal{S}} f_j(g(\boldsymbol{x}_i)))$.$\!$
In the \textbf{M-step}, the likelihood is maximized using posterior class probability from the previous E-step.
Theoretically, we show that our prototypical contrastive loss contributes to likelihood maximization by clustering similar instances, resulting in compact features.

In the corrupted generation stage, we assume a transition matrix $T_v\!\left(u\right)$, which means the probability of label $u$ transiting to label $v$, where $\sum_{v=1}^{C} T_v\!\left(u\right)=1$.
In the second stage that generating partial labels, we further make a general assumption \cite{LiuD12} that each label $y_i$ have the following probability 
$\sum_{u=1}^{C}\!T_{y_i}(u)\!\cdot\!\hbar\!\left(\mathcal{S}_i\right)$ of generating $\mathcal{S}$, where $\hbar(\cdot)$ is a function making it a valid probability distribution.

\subsection{Analyzing the E-step}
In \textbf{E-step}, we first introduce $q_i(c)$ as the density function of a possible distribution over $C$-class for instance $\boldsymbol{x}_i (1\!\le\!i \le\!n, 1\!\le c \!\le\!C)$. Let $\boldsymbol{\theta}$ be the total parameters of the encoder $g(\cdot)$ and the classifier $f(\cdot)$. Our goal is to maximize the likelihood below,
\begin{align*}
\label{eq:Estep}
   &\underset{\boldsymbol{\theta}}{\operatorname{argmax}} \sum_{i=1}^{n} \log p\!\left(\mathcal{S}_{i}, \boldsymbol{x}_{i}\!\mid\!\boldsymbol{\theta}\right) \geq \\
   &\underset{\boldsymbol{\theta}}{\operatorname{argmax}} \sum_{i=1}^{n} \sum_{y_{i}=1}^C q_i\!\left(y_i\right) \log \sum_{u=1}^{C}T_{y_i}\!(u) \!\cdot\! \frac{p\left(\boldsymbol{x}_{i}, y_{i}\!\mid\!\boldsymbol{\theta}\right)}{q_i\!\left(y_i\right)}.
\end{align*}

The proof for Eq. (10) can be found in Appendix A.1.
We assume that the transition matrix for generating noisy labels follows a \textit{general random distribution} which guarantees that each matrix column sums up to $\boldsymbol{\psi}$.
For clarity, we list some matrix representations of 5-classes as follows:
\begin{equation*}
\resizebox{0.8\linewidth}{!}{$
\left[\begin{array}{ccccc}
1-q_1 & q_1/4 & q_1/4 & q_1/4 & q_1/4 \\
q_1/4 & 1-q_1 & q_1/4 & q_1/4 & q_1/4 \\
q_1/4 & q_1/4 & 1-q_1 & q_1/4 & q_1/4 \\
q_1/4 & q_1/4 & q_1/4 & 1-q_1 & q_1/4 \\
q_1/4 & q_1/4 & q_1/4 & q_1/4 & 1-q_1
\end{array}\right],
\left[\begin{array}{ccccc}
q_1 & q_2 & q_3 & q_4 & q_5 \\
q_2 & q_3 & q_4 & q_5 & q_1 \\
q_3 & q_4 & q_5 & q_1 & q_2 \\
q_4 & q_5 & q_1 & q_2 & q_3\\
q_5 & q_1 & q_2 & q_3 & q_4
\end{array}\right],$}
\end{equation*}
where the first one is uniform random distribution, and the second one is a \textit{general random distribution} which takes the uniform noise as a special case. 
By leveraging the concavity of the $\log(\cdot)$ function, the inequality holds with equality when $\sum_{u=1}^{C}\!T_{y_i}\!\left(u\right)\!\cdot\!\frac{p\left(\boldsymbol{x}_{i}, y_{i} \mid \boldsymbol{\theta}\right)}{q_i\left(y_i\right)}$ is a constant.
Then we can get $q_i\!\left(y_i\right)\!=\!p\!\left(y_{i}\!\mid\!\boldsymbol{x}_{i}, \!\boldsymbol{\theta}\right)$ as the posterior class probability, whose proof can be found in Appendix A.2.
Since each example belongs to exactly one cluster, we set $q_i(j)\!=\!1$ for $j\!=\!\text{argmax}_{j \in \mathcal{S}}f_j(g(\boldsymbol{x}_i))$ and $q_i(j)\!=\!0$ elsewise.

\subsection{Analyzing the M-step}
In \textbf{M-step}, after getting $q_i(y_i)$ in E-step, the likelihood maximization is optimized by
\begin{equation}
\label{eq:likelihood}
    \underset{\boldsymbol{\theta}}{\operatorname{argmax}} \sum_{i=1}^{n} \sum_{y_{i}=1}^C q_i\!\left(y_i\right) \log\left( \boldsymbol{\psi} \!\cdot\! \frac{p\left(\boldsymbol{x}_{i}, y_{i}\!\mid\!\boldsymbol{\theta}\right)}{q_i\!\left(y_i\right)}\right).
\end{equation}
To achieve the optimization target of Eq. (\ref{eq:likelihood}), we rewrite it according to Lemma \ref{lemma1}.

\newtheorem{lemma}{\bf Lemma}
\begin{lemma}\label{lemma1}
We assume the normalized embeddings in the hyperspherical space follow the von Mises-Fisher (vMF) distribution \cite{fisher1953dispersion}, whose probabilistic density is given by $f\left(x \mid \bar{\delta}_{i}, \kappa\right)=c_{d}(\kappa) e^{\kappa \overline{\boldsymbol{\delta}}_{i}^{\top} g(\boldsymbol{x})}$, where $\overline{\boldsymbol{\delta}}_{i}=\boldsymbol{\delta}_{i} /\left\|\boldsymbol{\delta}_{i}\right\|$ is the mean direction.
We define the set of instances with the same prediction $H(c)=\{\boldsymbol{x}_i \in \mathcal{D}|\hat{y}_i=c\}$. 
The maximization step can be explained by aligning the feature vector to the corresponding mean center $\boldsymbol{\delta}_c$:
\begin{align*}
&\underset{\boldsymbol{\theta}}{\operatorname{argmax}} \sum_{i=1}^{n} \sum_{y_{i}=1}^C q_i\!\left(y_i\right) \log\left( \boldsymbol{\psi} \!\cdot\! \frac{p\left(\boldsymbol{x}_{i}, y_{i}\!\mid\!\boldsymbol{\theta}\right)}{q_i\!\left(y_i\right)}\right)\\
&= \underset{\theta}{\operatorname{argmin}} \sum_{y_{i}=1}^C \sum_{\boldsymbol{x} \in H(y_i)} \left\|g(\boldsymbol{x})-\boldsymbol{\delta}_{y_{i}}\right\|^{2}.
\end{align*}
\end{lemma}
The proof of Lemma \ref{lemma1} can be found in Appendix A.3. 
We then show that minimizing the prototypical contrastive loss is approximately maximizing the likelihood. 
We can decompose the loss as follows:
\begin{align*}
\mathcal{L}_{\mathrm{pc}}(g;\mathcal{D},\boldsymbol{p},\boldsymbol{z}) &=-\frac{1}{n}\!\sum_{\boldsymbol{x} \in \mathcal{D}}\log\frac{\exp(g(\boldsymbol{x})\!\cdot\!\boldsymbol{p} / \tau)}
{\sum_{c=1}^{C}\exp(g(\boldsymbol{x})\!\cdot\!\boldsymbol{z}_{c}/\tau)}\\
&=\frac{1}{n}\!\sum_{\boldsymbol{x} \in \mathcal{D}}\!\left\{-g\!\left(\boldsymbol{x}\right)\!\cdot\!\boldsymbol{p} / \tau \right\}+\mathcal{R}(g;\mathcal{D},\boldsymbol{z}),
\end{align*}
where the first term refers to the alignment term \cite{wang2020understanding}, which encourages the compactness of features for positive pairs, and the second term $\mathcal{R}(g;\mathcal{D},\boldsymbol{z})$ is the uniformity term that benefits information-preserving \cite{wang2020understanding}. To see this, we have the following lemma \ref{lemma2} proved in Appendix A.4.

\begin{lemma}\label{lemma2}
Minimizing the alignment term can be explained to align the feature vector to the corresponding mean center $\boldsymbol{\delta}_c$.
\begin{align*}
&\underset{\theta}{\operatorname{argmin}} \frac{1}{n} \sum_{\boldsymbol{x} \in \mathcal{D}}\!\left\{-g\!\left(\boldsymbol{x}\right)\!\cdot\!\boldsymbol{p} / \tau \right\}\\
&=\underset{\boldsymbol{\theta}}{\operatorname{argmin}} \sum_{y_{i}=1}^C \sum_{\boldsymbol{x} \in H(y_i)} \left\|g(\boldsymbol{x})-\boldsymbol{\delta}_{y_{i}}\right\|^{2}.
\end{align*}
\end{lemma}

\section{Conclusion}
This research delves into the realm of Unreliable Partial Label Learning, focusing particularly on the challenges presented by unreliable and ambiguous labels. Through our novel method, we incorporate Unreliability-Robust Contrastive Learning into the learning process, making the model more resistant to unreliablity and enhancing overall learning efficacy. Additionally, our innovative strategy of KNN-based candidate label set correction and consistency regularization-based label disambiguation assists in refining label quality and improving representation. These proposed techniques, as evidenced by our experimental results across diverse datasets, demonstrate significant advancements in handling unreliability and partiality in labels, thereby leading to state-of-the-art performance. Furthermore, our theoretical analysis from the Expectation Maximization algorithm perspective provides a deeper understanding of our method's effectiveness. 

\bibliography{egbib}

\appendix
\onecolumn

\section{A.\quad Proof Details}
\subsection{A.1\quad Proof of Eq. (10).}
\begin{equation*}
\begin{split}
&\!\!\!\!\!\!\underset{\theta}{\operatorname{argmax}} \sum_{i=1}^{n} \log p\!\left(\mathcal{S}_{i}, \boldsymbol{x}_{i}\!\mid\!\theta\right) \\
&\!\!\!\!\!\!=\underset{\theta}{\operatorname{argmax}} \sum_{i=1}^{n} \log \sum_{y_i=1}^C  p\!\left(\mathcal{S}_i|y_i\right)\!\cdot\! p\!\left(\boldsymbol{x}_{i},\!y_{i}\!\mid\!\theta\right) \\
&\!\!\!\!\!\!=\underset{\theta}{\operatorname{argmax}} \sum_{i=1}^{n} \log \sum_{y_i=1}^C  \sum_{u=1}^{C}T_{y_i}\!(u) \!\cdot\! \hbar\left(\mathcal{S}_i\right)\!\cdot\! p\!\left(\boldsymbol{x}_{i},\!y_{i}\!\mid\!\theta\right) \\
&\!\!\!\!\!\!=\underset{\theta}{\operatorname{argmax}} \sum_{i=1}^{n} \log \sum_{y_i=1}^C  \sum_{u=1}^{C}T_{y_i}\!(u) \!\cdot\! p\!\left(\boldsymbol{x}_{i},\!y_{i}\!\mid\!\theta\right)+\sum_{i=1}^{n} \log \left(\hbar\left(\mathcal{S}_{i}\right)\right) \\
&\!\!\!\!\!\!=\underset{\theta}{\operatorname{argmax}} \sum_{i=1}^{n} \log \sum_{y_{i}=1}^C \sum_{u=1}^{C}T_{y_i}\!(u) \!\cdot\! q_i\!\left(y_i\right) \frac{p\left(\boldsymbol{x}_{i}, y_{i}\!\mid\!\theta\right)}{q_i\!\left(y_i\right)} \\
&\!\!\!\!\!\!\geq \underset{\theta}{\operatorname{argmax}} \sum_{i=1}^{n} \sum_{y_{i}=1}^C q_i\!\left(y_i\right) \log \sum_{u=1}^{C}T_{y_i}\!(u) \!\cdot\! \frac{p\left(\boldsymbol{x}_{i}, y_{i}\!\mid\!\theta\right)}{q_i\!\left(y_i\right)} .
\end{split}
\end{equation*}

\subsection{A.2\quad Proof of 
$q_i\!\left(y_i\right)=p\!\left(y_{i}\!\mid\!\boldsymbol{x}_{i}, \theta\right)$.}
With the assumption that the transition matrix in the generation of noisy label follows a \textit{general random distribution}, which guarantees that each matrix column sums up to $\boldsymbol{\psi}$.
For clarity, we list some matrix representations of 5-classes noise generations as follows:
\begin{equation*}
\left[\begin{array}{ccccc}
1-q_1 & q_1/4 & q_1/4 & q_1/4 & q_1/4 \\
q_1/4 & 1-q_1 & q_1/4 & q_1/4 & q_1/4 \\
q_1/4 & q_1/4 & 1-q_1 & q_1/4 & q_1/4 \\
q_1/4 & q_1/4 & q_1/4 & 1-q_1 & q_1/4 \\
q_1/4 & q_1/4 & q_1/4 & q_1/4 & 1-q_1
\end{array}\right],
\left[\begin{array}{ccccc}
q_1 & q_2 & q_3 & q_4 & q_5 \\
q_2 & q_3 & q_4 & q_5 & q_1 \\
q_3 & q_4 & q_5 & q_1 & q_2 \\
q_4 & q_5 & q_1 & q_2 & q_3\\
q_5 & q_1 & q_2 & q_3 & q_4
\end{array}\right],
\end{equation*}
where the first one is uniform random distribution, and the second one is a \textit{general random distribution} which takes the uniform noise as a special case. 

By using the fact that $\log(\cdot)$ function is concave, the inequality holds with equality when $\sum_{u=1}^{C}T_{y_i}\!\left(u\right)\!\cdot\!\frac{p\left(\boldsymbol{x}_{i}, y_{i} \mid \theta\right)}{q_i\left(y_i\right)}$ is a constant $Z$.
We integrate $y_i$ on both sides simultaneously, 
\begin{equation*}
\begin{gathered}
q_i\!\left(y_i\right)=\frac{1}{Z} \sum_{u=1}^{C}T_{y_i}\!(u) 
\!\cdot\! p\!\left(\boldsymbol{x}_{i}, y_{i}\!\mid\!\theta\right),\\
\int_{y_i}\!q_i\!\left(y_i\right) \mathrm{d}{y_i}=\int_{y_i}\! \frac{1}{Z} \sum_{u=1}^{C}T_{y_i}\!(u) 
\!\cdot\! p\!\left(\boldsymbol{x}_{i}, y_{i}\!\mid\!\theta\right) \mathrm{d}{y_i}=\frac{\boldsymbol{\psi}}{Z}\!\!\int_{y_i}\! 
\!\cdot\! p\!\left(\boldsymbol{x}_{i}, y_{i}\!\mid\!\theta\right) \mathrm{d}{y_i}=\frac{\boldsymbol{\psi}}{Z} p(\boldsymbol{x}_i|\theta)=1,\\
q_i\!\left(y_i\right)=\frac{\boldsymbol{\psi}}{Z} p\!\left(\boldsymbol{x}_{i}, y_{i}\!\mid\!\theta\right)=\frac{p\!\left(\boldsymbol{x}_{i}, y_{i}\!\mid\!\theta\right)}{p(\boldsymbol{x}_i|\theta)}=p\!\left(y_{i}\!\mid\!\boldsymbol{x}_{i}, \theta\right)   ,
\end{gathered}
\end{equation*}
where $p\!\left(y_{i}\!\mid\!\boldsymbol{x}_{i}\right)$ is the posterior class probability.

\subsection{A.3\quad Proof of Lemma 1.}
Here, we give a detailed proof of Lemma 1. 
Firstly, the probabilistic density of the von Mises-Fisher (vMF) distribution \cite{fisher1953dispersion} is given by $f\left(x \mid \bar{\delta}_{i}, \kappa\right)=c_{d}(\kappa) e^{\kappa \overline{\boldsymbol{\delta}}_{i}^{\top} g(\boldsymbol{x})}$, where $\overline{\boldsymbol{\delta}}_{i}=\boldsymbol{\delta}_{i} /\left\|\boldsymbol{\delta}_{i}\right\|$ is the mean direction, $\kappa$ is the concentration parameter, and $c_d(\kappa)$ is the normalization factor.

The likelihood can be simplification under the von Mises-Fisher (vMF) distribution as follows:
\begin{equation*}
\begin{split}
&\underset{\theta}{\operatorname{argmax}} \sum_{i=1}^{n} \sum_{y_{i}=1}^C q_i\!\left(y_i\right) \log\left( \boldsymbol{\psi} \!\cdot\! \frac{p\left(\boldsymbol{x}_{i}, y_{i}\!\mid\!\theta\right)}{q_i\!\left(y_i\right)}\right)\\
&=\underset{\theta}{\operatorname{argmax}} \sum_{i=1}^{n} \sum_{y_{i}=1}^C q_i\!\left(y_i\right) \log p\left(\boldsymbol{x}_{i}\!\mid\!\theta, y_{i}\right)p\left( y_{i}\right)\\
&=\underset{\theta}{\operatorname{argmax}} \sum_{i=1}^{n} \sum_{y_{i}=1}^C \mathbb{I}(\hat{y}_i\!=\!j) \log p\left(\boldsymbol{x}_{i}\!\mid\!\theta, y_{i}\right)\\
&=\underset{\theta}{\operatorname{argmax}} \sum_{y_{i}=1}^C \sum_{\boldsymbol{x}_{i} \in H(y_i)} \log p\left(\boldsymbol{x}_{i}\!\mid\!\theta, y_{i}\right)\\
&=\underset{\theta}{\operatorname{argmax}} \sum_{y_{i}=1}^C \sum_{\boldsymbol{x} \in H(y_i)} \left(\kappa \overline{\boldsymbol{\delta}}_{y_i}^{\top} g(\boldsymbol{x})+\log \left(c_{d}(\kappa)\right)\right)\\
&=\underset{\theta}{\operatorname{argmax}} \sum_{y_{i}=1}^C \frac{n_{y_i}}{n} \left \|\boldsymbol{\delta}_{y_i} \right \|,
\end{split}
\end{equation*}
where $n_{y_i}=|H(y_i)|, \boldsymbol{\delta}_{y_i}=\frac{1}{n_{y_i}} \sum_{\boldsymbol{x} \in H(y_i)} g(\boldsymbol{x})$.

The equation on the right side of the lemma 1 can be reduced to as follows:
\begin{equation*}
\begin{split}
&\underset{\theta}{\operatorname{argmin}} \sum_{y_{i}=1}^C \sum_{\boldsymbol{x} \in H(y_i)} \left\|g(\boldsymbol{x})-\boldsymbol{\delta}_{y_i}\right\|^{2}\\
&=\underset{\theta}{\operatorname{argmin}} \sum_{y_{i}=1}^C \sum_{\boldsymbol{x} \in H(y_i)} \left( 
\left\|g(\boldsymbol{x})\right\|^{2}-2 g(\boldsymbol{x})^{\top} \boldsymbol{\delta}_{y_i}+\left\|\boldsymbol{\delta}_{y_i}\right\|^{2}\right)\\
&=\underset{\theta}{\operatorname{argmin}} \sum_{y_{i}=1}^C \left(n_{y_i}-n_{y_i}\left\|\boldsymbol{\delta}_{y_i}\right\|^{2} \right)\\
&=\underset{\theta}{\operatorname{argmax}} \sum_{y_{i}=1}^C \frac{n_{y_i}}{n} \left \|\boldsymbol{\delta}_{y_i} \right \|^2.
\end{split}
\end{equation*}

Note that the contrastive embeddings are distributed on the hypersphere and thus $\left\|\boldsymbol{\delta}_j\right\| \in [0,1]$. It can be directly derived that,
\begin{equation*}
\sum_{y_{i}=1}^C \frac{n_{y_i}}{n} \left \|\boldsymbol{\mu}_{y_i} \right \| \ge \sum_{y_{i}=1}^C \frac{n_{y_i}}{n} \left \|\boldsymbol{\mu}_{y_i} \right \|^2
\end{equation*}

Therefore, maximizing the alignment loss is equivalent to maximizing a lower bound of the likelihood.

\subsection{A.4\quad Proof of Lemma 2.}
\begin{equation*}
\begin{split}
&\underset{\theta}{\operatorname{argmin}} \frac{1}{n} \sum_{\boldsymbol{x} \in \mathcal{D}}\!\left\{-g\!\left(\boldsymbol{x}\right)\!\cdot\!\boldsymbol{p} / \tau \right\}\\
& =\frac{1}{n} \sum_{\boldsymbol{x} \in \mathcal{D}} \left(\left\|g\!\left(\boldsymbol{x}\right)-\boldsymbol{p}\right\|^{2}-2\right) /(2 \tau)\\
& =\frac{1}{2 \tau n}  \sum_{\boldsymbol{x} \in \mathcal{D}}\left\|g(\boldsymbol{x})-\boldsymbol{p}\right\|^{2}+K\\
& =\frac{1}{2 \tau n} \sum_{y_{i}=1}^C \sum_{\boldsymbol{x} \in H(y_i)}\left\|g(\boldsymbol{x})-\boldsymbol{p}\right\|^{2}+K,
\end{split}
\end{equation*}
where $K$ is a constant.As model training, $l$ gradually becomes a one-hot probability vectors, where $\boldsymbol{p}=\sum_{c=1}^{C} \boldsymbol{z}_c\cdot l_c \approx \boldsymbol{z}_{y_i}=\frac{1}{n_{y_i}} \sum_{\boldsymbol{x} \in H(y_i)} g(\boldsymbol{x})=\boldsymbol{\delta_{y_i}}$.
Therefore, minimizing the alignment term can be explained to align the feature vector to the corresponding mean center $\boldsymbol{\delta}_c$.

\section{B.\quad Extended Experiments}

\subsection{B.1\quad Extended Implementation Details.}
Datasets:
\begin{itemize}
\item CIFAR-10 comprises 60,000 RGB images ($32 \times 32 \times 3$), distributed between 10 classes such as airplane, bird, automobile, cat, deer, frog, dog, horse, ship, and truck. The dataset is divided into 50,000 training and 10,000 test images, each class represented evenly.

\item CIFAR-100, similar to CIFAR-10, consists of 60,000 RGB images ($32 \times 32 \times 3$), albeit spanning 100 distinct classes. The dataset maintains a uniform distribution with 500 training images and 100 test images per class.

\item Fashion-MNIST (F-MNIST) provides a collection of 70,000 greyscale images ($28 \times 28 \times 1$) representing 10 different fashion items, including T-shirt/top, trouser, pullover, dress, sandal, coat, shirt, sneaker, bag, and ankle boot.
\end{itemize}

To ensure an unbiased comparison, we propose that the test accuracy should be reported when the validation set accuracy achieves its maximum. However, the above datasets do not inherently possess a validation set. To address this, we propose dividing the datasets into training, validation, and test sets in a 4:1:1 ratio, resulting in 40,000 training, 10,000 validation, and 10,000 test images for CIFAR-10 and CIFAR-100, and 46,666 training, 11,666 validation, and 11,668 test images for F-MNIST.

The CIFAR-10 and CIFAR-100 datasets undergo the following data augmentation techniques: (1) Random Horizontal Flipping, (2) Random Cropping, (3) Cutout, and (4) Auto Augment. Meanwhile, the F-MNIST dataset is subjected to (1) Random Horizontal Flipping, (2) Random Cropping, and (3) Cutout.
\subsection{B.2\quad Experiments on F-MNIST.}
The results presented in Table \ref{tab:fashion} accentuate the remarkable performance of our proposed Unreliability-Robust Representation Learning (URRL) method on the F-MNIST synthesized dataset. Regardless of the varying levels of unreliability ($\mu$) and partial rates ($\eta$), our method consistently yields the highest test accuracy, outperforming other comparative models such as RABS, PiCO, CR-DPLL, PRODEN, RC, and CC.

For instance, in scenarios of lower unreliability and partial rate (i.e., $\mu=0.1$, $\eta=0.1$), URRL exhibits robustness and high precision with an accuracy of $94.61\pm 0.13\%$, which notably surpasses the performance of other methods. Even in situations with increased levels of unreliability and partiality (i.e., $\mu=0.5$, $\eta=0.5$), our model continues to show superior performance, delivering an accuracy of $90.55\pm 0.35\%$. It is important to highlight that this performance holds up even when compared to models such as PiCO and CR-DPLL, which maintain relatively higher accuracies among the rest.

The consistent outperformance of URRL across various conditions demonstrates its effectiveness and robustness in learning from unreliable partial labels, proving its suitability for real-world scenarios where data uncertainty and ambiguity prevail.

\begin{table}[tbp]
    \centering
    \caption{Test accuracy (mean$\pm$std) on the F-MNIST synthesized dataset. Results demonstrating superior performance are distinguished by boldface.}
    \resizebox{1\columnwidth}{!}{
        \begin{tabular}{c|c|c|ccccccc}
            \midrule
            Dataset & \multicolumn{1}{c}{$\eta$} 
                & $\mu$         & Ours                        & RABS               & PiCO               & CR-DPLL            & PRODEN             & RC                 & CC                  \\
            \midrule
            \multirow{9}[6]{*}{F-MNIST} 
                & 0.1  & 0.1   & \textbf{94.61} $\pm$ 0.13\% & 92.51 $\pm$ 0.22\% & 94.08 $\pm$ 0.13\% & 94.10 $\pm$ 0.09\% & 91.38 $\pm$ 0.07\% & 90.96 $\pm$ 0.56\% & 90.98 $\pm$ 0.27\%  \\
                & 0.1  & 0.3   & \textbf{93.82} $\pm$ 0.17\% & 90.69 $\pm$ 0.33\% & 92.56 $\pm$ 0.18\% & 92.79 $\pm$ 0.21\% & 86.49 $\pm$ 0.42\% & 84.39 $\pm$ 2.05\% & 84.45 $\pm$ 0.08\%  \\
                & 0.1  & 0.5   & \textbf{92.31} $\pm$ 0.18\% & 86.41 $\pm$ 0.53\% & 90.20 $\pm$ 0.44\% & 91.11 $\pm$ 0.33\% & 81.13 $\pm$ 0.70\% & 78.60 $\pm$ 2.68\% & 79.02 $\pm$ 1.37\%  \\
                \cmidrule{2-10}      
                & 0.3  & 0.1   & \textbf{94.42} $\pm$ 0.07\% & 80.16 $\pm$ 1.73\% & 93.47 $\pm$ 0.23\% & 93.66 $\pm$ 0.24\% & 90.66 $\pm$ 0.28\% & 89.95 $\pm$ 0.28\% & 90.49 $\pm$ 0.79\%  \\
                & 0.3  & 0.3   & \textbf{93.17} $\pm$ 0.19\% & 76.61 $\pm$ 2.13\% & 91.69 $\pm$ 0.19\% & 92.58 $\pm$ 0.30\% & 86.31 $\pm$ 0.61\% & 84.53 $\pm$ 0.80\% & 84.92 $\pm$ 1.00\%  \\
                & 0.3  & 0.5   & \textbf{91.70} $\pm$ 0.22\% & 62.83 $\pm$ 1.77\% & 88.89 $\pm$ 0.18\% & 90.72 $\pm$ 0.26\% & 78.09 $\pm$ 2.28\% & 79.32 $\pm$ 1.56\% & 76.44 $\pm$ 1.83\% \\
                \cmidrule{2-10}      
                & 0.5  & 0.1   & \textbf{93.91} $\pm$ 0.17\% & 60.17 $\pm$ 3.06\% & 92.51 $\pm$ 0.09\% & 93.16 $\pm$ 0.13\% & 89.73 $\pm$ 0.30\% & 88.24 $\pm$ 0.28\% & 88.54 $\pm$ 0.57\%  \\
                & 0.5  & 0.3   & \textbf{92.92} $\pm$ 0.07\% & 53.13 $\pm$ 1.71\% & 89.52 $\pm$ 0.12\% & 91.30 $\pm$ 0.19\% & 85.32 $\pm$ 0.10\% & 83.27 $\pm$ 0.02\% & 81.50 $\pm$ 1.18\%  \\
                & 0.5  & 0.5   & \textbf{90.55} $\pm$ 0.35\% & 37.05 $\pm$ 5.45\% & 83.98 $\pm$ 0.36\% & 88.76 $\pm$ 0.40\% & 78.37 $\pm$ 0.63\% & 73.84 $\pm$ 0.31\% & 67.70 $\pm$ 5.80\%  \\
            \midrule
            \midrule
            \end{tabular}
        }
    \label{tab:fashion}
\end{table}

\subsection{B.3\quad Visual Analysis via T-SNE.}
\begin{figure}[!tbp]
  \centering
  \begin{subfigure}[b]{0.45\textwidth}
    \centering
    \includegraphics[width=\textwidth]{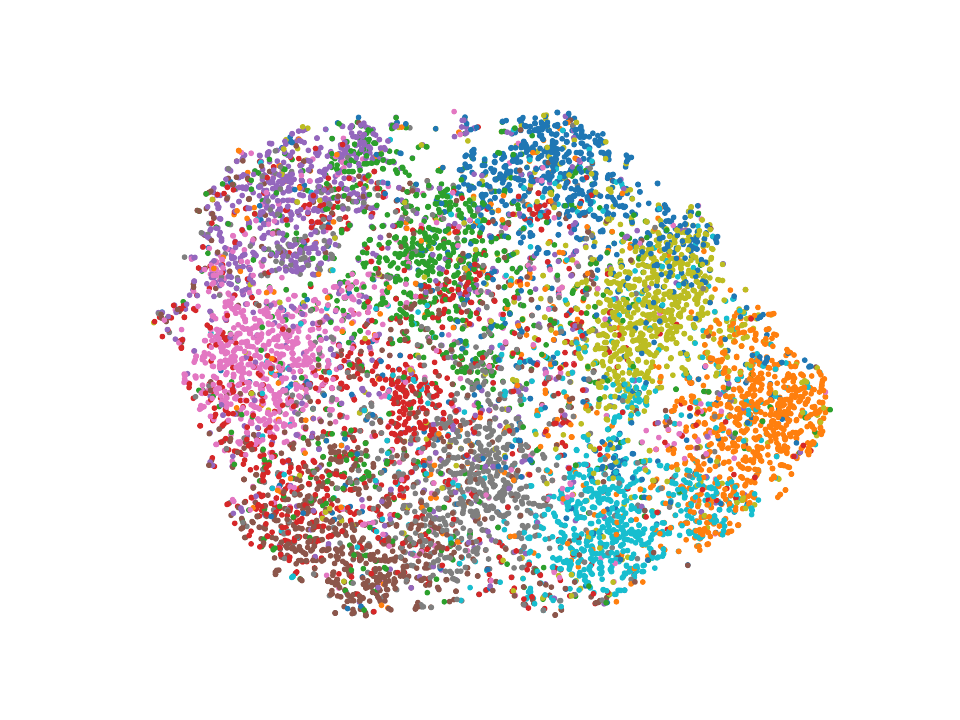}
    \caption{PRODEN}
    \label{fig:tsne1}
  \end{subfigure}
  \hfill
  \begin{subfigure}[b]{0.45\textwidth}
    \centering
    \includegraphics[width=\textwidth]{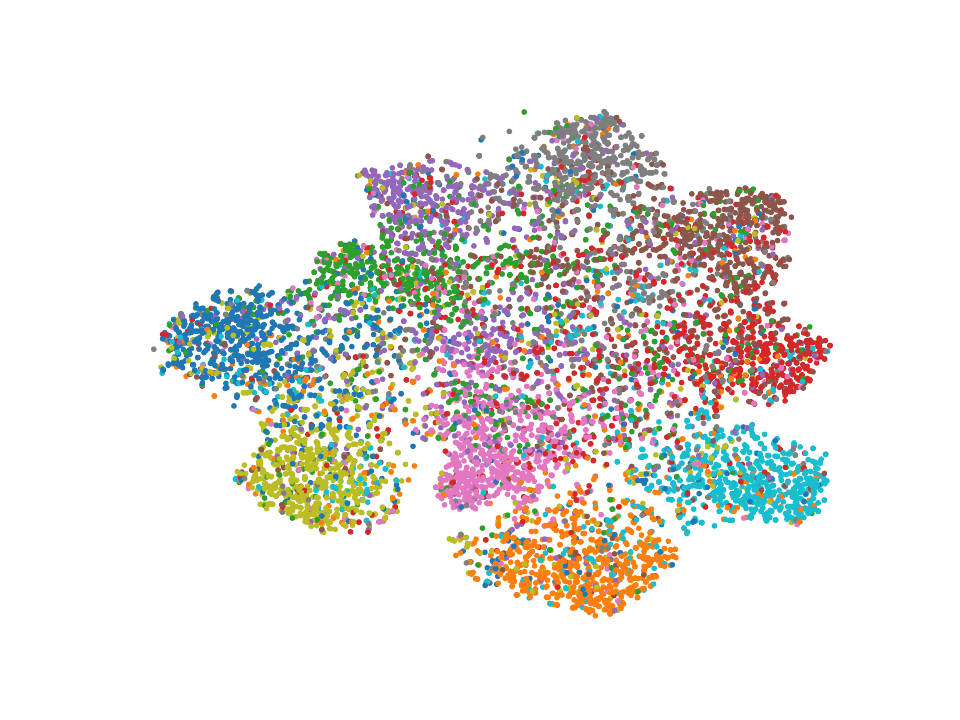}
    \caption{CC}
    \label{fig:tsne2}
  \end{subfigure}
  \\
  \begin{subfigure}[b]{0.45\textwidth}
    \centering
    \includegraphics[width=\textwidth]{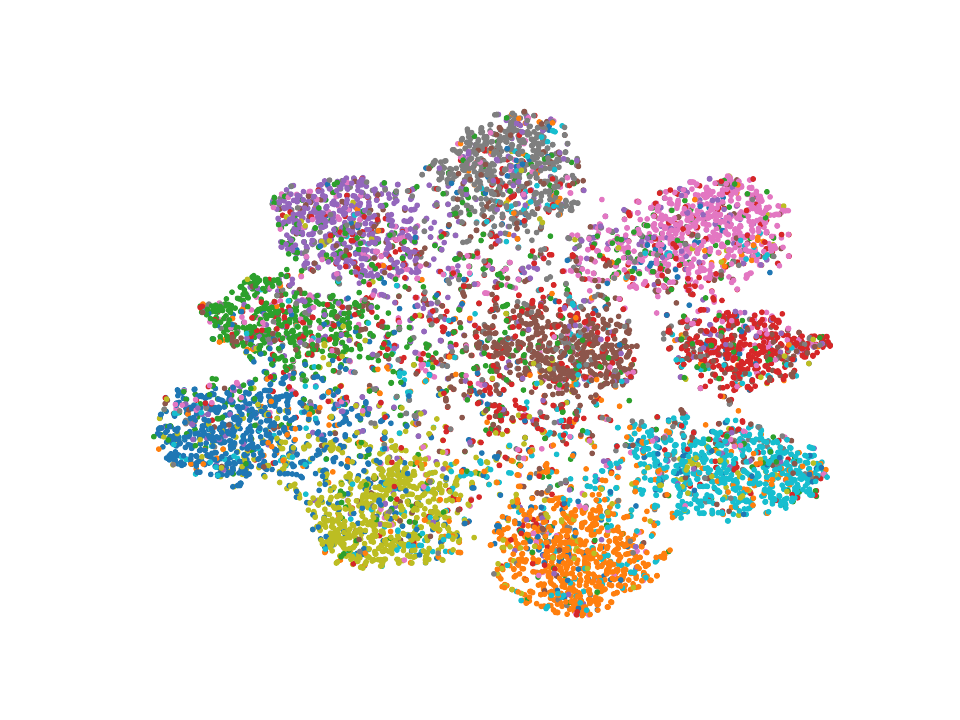}
    \caption{RC}
    \label{fig:tsne3}
  \end{subfigure}
  \hfill
  \begin{subfigure}[b]{0.45\textwidth}
    \centering
    \includegraphics[width=\textwidth]{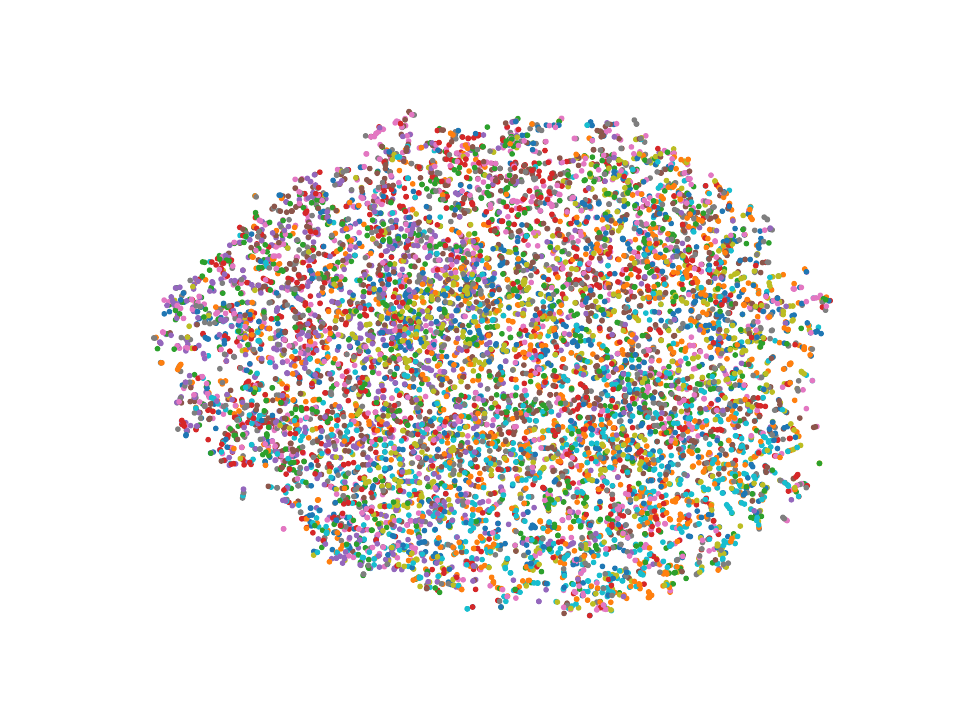}
    \caption{RABS}
    \label{fig:tsne4}
  \end{subfigure}
  \\
  \begin{subfigure}[b]{0.45\textwidth}
    \centering
    \includegraphics[width=\textwidth]{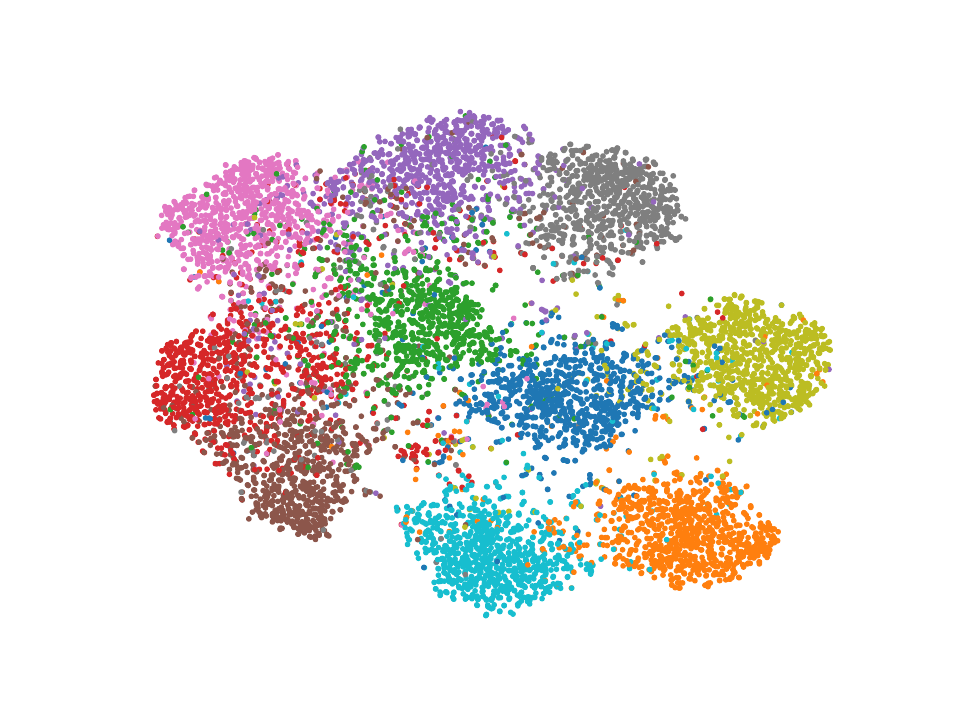}
    \caption{PiCO}
    \label{fig:tsne5}
  \end{subfigure}
  \hfill
  \begin{subfigure}[b]{0.45\textwidth}
    \centering
    \includegraphics[width=\textwidth]{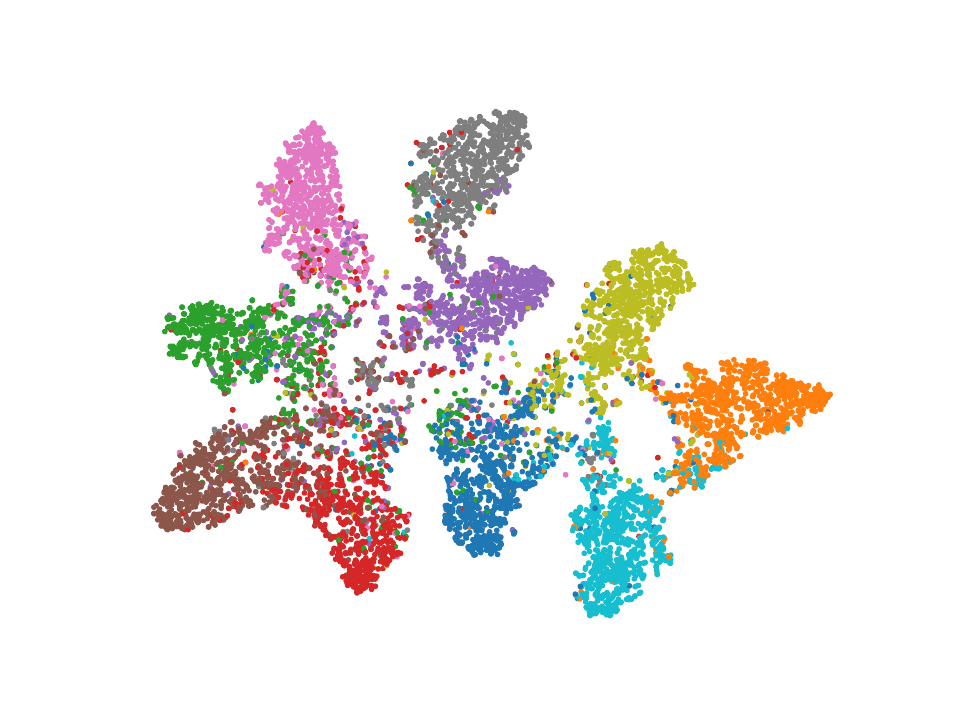}
    \caption{CR-DPLL}
    \label{fig:tsne6}
  \end{subfigure}
  \\
  \begin{subfigure}[b]{0.45\textwidth}
    \centering
    \includegraphics[width=\textwidth]{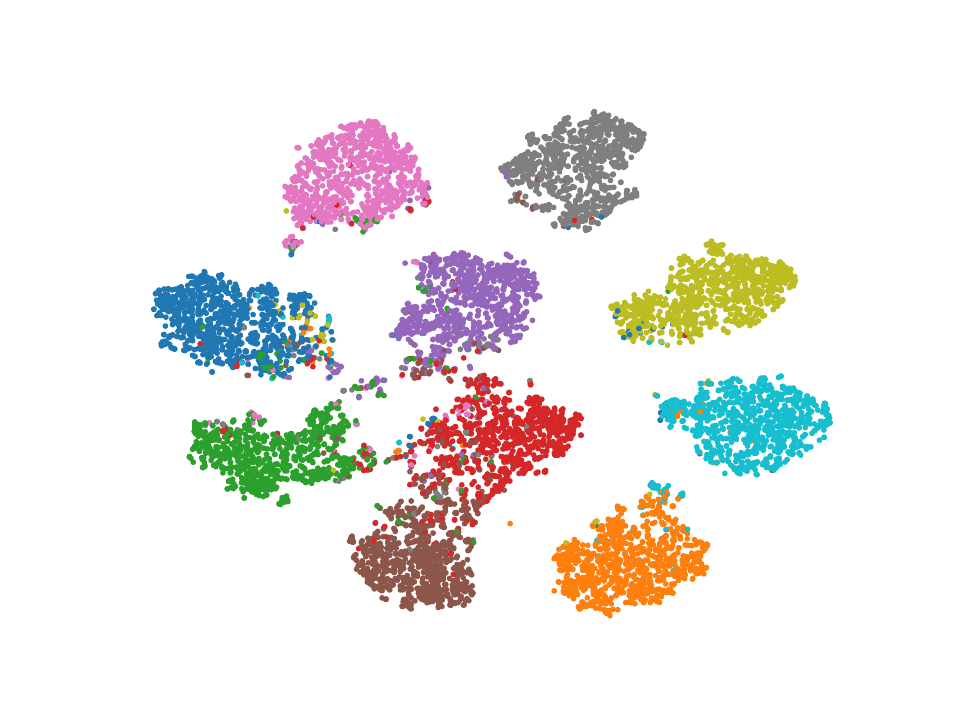}
    \caption{Ours}
    \label{fig:tsne7}
  \end{subfigure}

  \caption{Visualization of image representations generated by the feature encoder, as portrayed using t-distributed Stochastic Neighbor Embedding (t-SNE). Different colors denote different ground-truth class labels. The plot contrasts the t-SNE embeddings of six baseline methods (RC, PRODEN, CC, RABS, and PiCO, CR-DPLL) with our proposed URRL method on the CIFAR-10 dataset under the parameters $\eta=0.3$ and $\mu=0.3$.}
  \label{fig:tsne}
\end{figure}

We employ the t-distributed Stochastic Neighbor Embedding (t-SNE) technique \cite{van2008visualizing} to visualize the image representations generated by our feature encoder, as shown in Figure \ref{fig:tsne}. Different colors are used to distinguish among different ground-truth class labels. For this analysis, we utilize the CIFAR-10 dataset with parameters $\eta=0.3, \mu=0.3$. Our URRL method is contrasted with six benchmark baselines.

As can be observed, the t-SNE embeddings of the RC, PRODEN, CC, RABS, and PiCO models are largely indistinguishable, demonstrating the struggles these models face due to high levels of unreliability in their supervision signals. Meanwhile, the CR-DPLL model exhibits improved features, but still suffers from class overlap (for instance, between brown and red).

In stark contrast, the URRL method yields well-separated clusters and far more distinguishable representations, underscoring its effectiveness in learning high-quality representations. This indicates a clear advantage of URRL in dealing with unreliable partial label learning.

\end{document}